\documentclass[letterpaper]{article} 
\usepackage{aaai23}  
\usepackage{times}  
\usepackage{helvet}  
\usepackage{courier}  
\usepackage[hyphens]{url}  
\usepackage{graphicx} 
\usepackage{natbib}  
\usepackage{caption} 
\usepackage{algorithm}
\usepackage{algorithmic}
\usepackage{amsmath}
\usepackage{amsfonts}

\usepackage{newfloat}
\usepackage{listings}
\usepackage{multicol}
\usepackage{multirow}
\usepackage{makecell}
\usepackage{array}
\DeclareCaptionStyle{ruled}{labelfont=normalfont,labelsep=colon,strut=off} 
\floatstyle{ruled}
\newfloat{listing}{tb}{lst}{}
\floatname{listing}{Listing}
\title{Rethinking Alignment and Uniformity in Unsupervised Image Semantic Segmentation}
\author{
    Daoan Zhang\textsuperscript{\rm 1,\rm 3}, Chenming Li\textsuperscript{\rm 1}, Haoquan Li\textsuperscript{\rm 1}, Wenjian Huang\textsuperscript{\rm 1}, Lingyun Huang\textsuperscript{\rm 3}, Jianguo Zhang\textsuperscript{\rm 1, 2}\thanks{Corresponding Author}\\
}
\affiliations{
    \textsuperscript{\rm 1}Southern University of Science and Technology \\
    \textsuperscript{\rm 2}Peng Cheng Laboratory \\
    \textsuperscript{\rm 3}Ping An Technology (Shenzhen) Co., Ltd.\\
    \{12032503, 12132339, 12032492\}@mail.sustech.edu.cn, lingyun.huanguw@foxmail.com, \{huangwj, zhangjg\}@sustech.edu.cn

}

\usepackage{bibentry}

\begin{document}

\maketitle

\begin{abstract}
Unsupervised image semantic segmentation (UISS) aims to match low-level visual features with semantic-level representations without outer supervision. In this paper, we address the critical properties from the view of feature alignments and feature uniformity for UISS models. We also make a comparison between UISS and image-wise representation learning. Based on the analysis, we argue that the existing MI-based methods in UISS suffer from representation collapse. By this, we proposed a robust network called \textit{Semantic Attention Network (SAN)}, in which a new module \textit{Semantic Attention (SEAT)} is proposed to generate pixel-wise and semantic features dynamically. Experimental results on multiple semantic segmentation benchmarks show that our unsupervised segmentation framework specializes in catching semantic representations, which outperforms all the unpretrained and even several pretrained methods. 

\end{abstract}

\section{Introduction}
Semantic segmentation of images\cite{wang2023Dionysus} is to group pixels into regions with different semantic categories. Most deep-learning-based semantic segmentation approaches\cite{chen2018encoder}\cite{zhao2017pyramid, wang2023ftso}require myriad manual efforts to delineate images, especially for fine-grained segmentation tasks. To alleviate the demand for manual annotation, there is increasing attention to developing approaches\cite{wen2022den} to segment images in the \textit{absence} of image annotations, i.e., unsupervised image semantic segmentation. Compared to fully supervised segmentation or image-wise self-supervised learning, UISS is a much more complex problem. It aims to make \textit{dense} prediction between images without any annotations or heuristic information. Thus, the model can neither build the pixel-wise relations between images and annotations like fully supervised segmentation nor build image-wise relations like image-wise self-supervised learning. For UISS, we have to build relations between pixel-wise features and the generated semantic features. 

Besides, as the model has no supervision, we must find specific constraints to enable the model to learn good semantic representations. 

Accordingly, for UISS, The following two requirements should be satisfied: 

(1) The proposed method should build the relationship between pixels and semantics.

(2) We should design specific supervision for UISS to learn good semantics.

\begin{figure}[t]
\centering
\includegraphics[width=1\columnwidth]{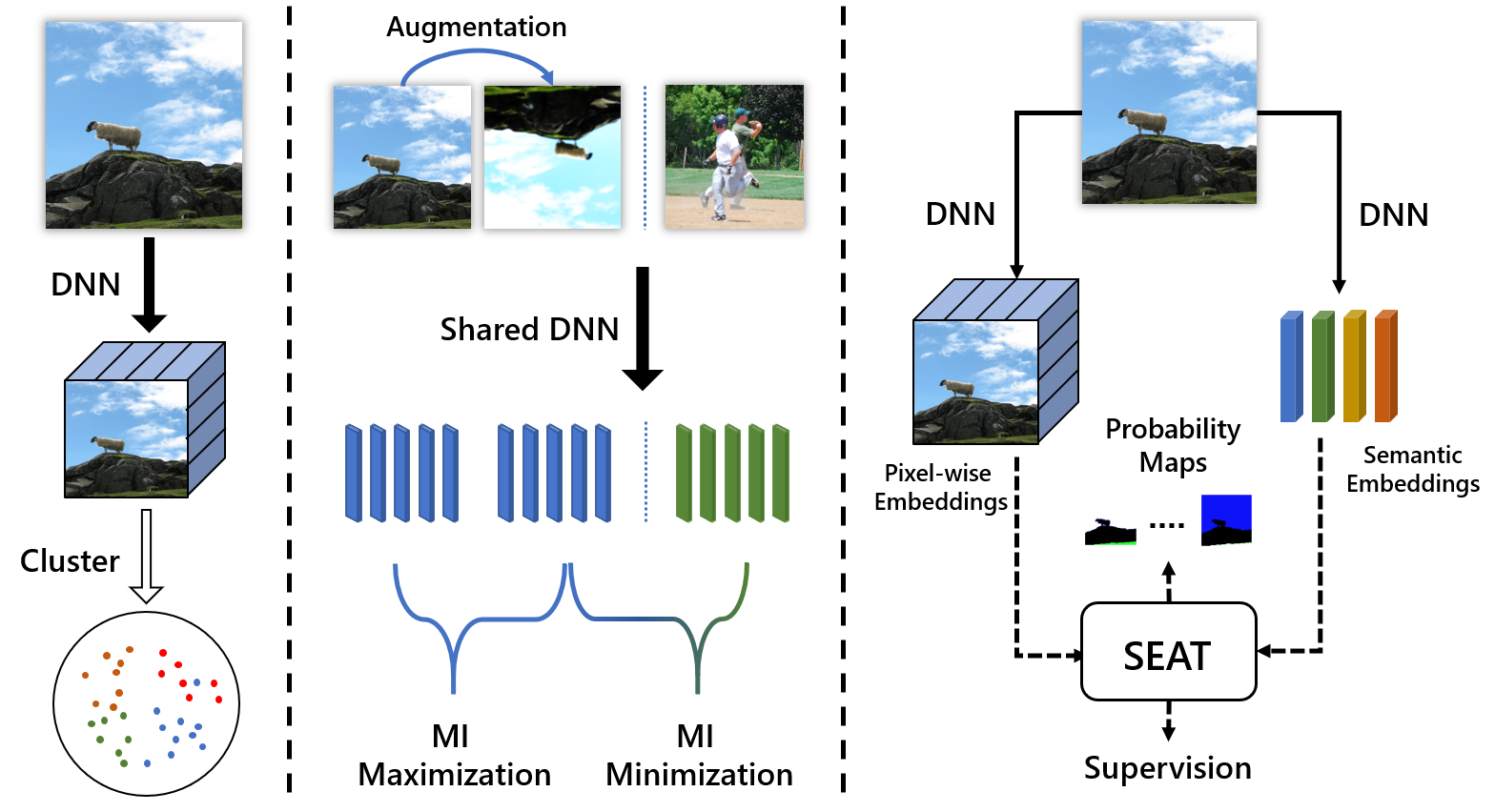} 
\caption{{ Paradigms of Unsupervised Semantic Segmentation.} { Clustering methods}(left) map the image into latent space and use cluster to classify semantics embeddings;
  { Contrastive methods}(middle) compare MI of different views from different images;
  { SAN} (right) maps both pixel-wise and semantic embeddings into latent space, then match them via the SEAT module. Specific supervision is applied to the output of SEAT.}
\label{frame}
\end{figure}

To build the relations of pixels and semantics, traditional unsupervised segmentation Methods\cite{caron2018deep} \cite{van2021unsupervised} generate high-level semantic information across pixels to classify pixels into clusters (Fig. \ref{frame}. left). These methods group pixels with \textit{low-level} feature representations into a few \textit{semantic} groups, often with some manually designed priors such as different augmentations. The supervision for these models usually comes from the pseudo-labels generated by the clustering. Essentially, it assumes pixels in the same semantic object share similar low-level embedding (features) in the latent space. However, there is often no guarantee that the low-level features in the latent space could represent the high-level semantic information. Thus far, clustering-based methods lack a mechanism to reinforce the low-level features to represent semantic information without considering high-level semantic information. In a nutshell, these methods do not meet the conditions in the requirement (2).

To address this problem, recent methods\cite{hjelm2018learning}\cite{cho2021picie}\cite{ji2019invariant} utilize mutual information(MI) as a supervision to generate high-level semantic information. The most well-known methods are based on contrastive learning, as shown in the middle of Fig. \ref{frame}, where models are trained by increasing the distance between representations from different images while reinforcing the similarity between the representations of different augmented images from the same image. However, such a design draws the attention of these models to focus on \textit{image-level} features or classification-friendly features, which may not be suitable for segmentation tasks, i.e., these models can miss \textit{fine-grained} information between pixels alone with structured semantic layouts. This preference also tends to cause the collapse of representations of unsupervised models in practice which appears to fail to classify pixels. That might be because contrastive methods only extract image-level representations from the whole dataset, which has little power in differentiating pixels.

In this study, to solve the problems of existing cluster-based and MI-based methods approaches, we propose a novel paradigm for UISS to build the relationship between pixels and semantics and can better generate accurate semantic features. 
When a batch of images is sent into the model to generate semantic information, the maximal information we can use is the complete batch-wise information. Therefore, We introduce an interactive module called \textit{Semantic Attention(SEAT)} to capture the relations between the pixel-wise and the batch-wise semantic information. Unlike the classical paradigms of cluster-based or MI-based methods, which only use one kind of information, our strategy uses both high(batch) and low(pixel) level information simultaneously so that the model can dynamically build the relationship between pixels and semantics(Fig.\ref{frame}. right). 

In SEAT, a mixture co-training structure is designed to generate pixel-wise and semantic-wise information via variant encoders and fuse them in the latent space via a shared decoder to learn a better representation of the whole image. To better generate global-wise semantic representations, ViT backbone is utilized to generate semantic embeddings due to the strong ability of ViT to extract global information. Meanwhile, the extraction of pixel-wise embeddings depends on CNN, which is sensitive to local and regional information. Compared to other methods, our method can unify both pixel-wise and semantic-wise information to better deal with semantic segmentation. 

Regarding the supervision design issue, similar to the UISS task, image-wise representation learning also faces the lack of annotation problem. Most of the previous image-wise representation learning methods utilize contrastive learning as a solution, which can also be considered a kind of MI-based approach \cite{hjelm2018learning}. We bridge the gap between image-wise representation learning and pixel-wise representation learning(i.e., UISS) by analyzing the connections and discrepancies between the two fields.

For MI-based image-wise representation learning, \cite{wang2020understanding} proposed two fundamental properties: (1) alignment of features from positive pairs and (2) uniformity of the distribution of the semantic features on the hypersphere. We argue that the missing supervision of pixel-wise embeddings is the culprit of the model class. Thus for pixel-wise representation learning, the properties should be: (1) alignment of features between pixels and semantics and (2) uniformity of the distribution of the semantic and pixel-wise features; because directly utilizing MI-based methods in UISS is proved to collapse easily. Therefore, we introduce the image reconstruction task instead of the MI-based methods, which can prevent the model 
from getting invalid. The structural information can be learned implicitly via our image reconstruction approach (see Fig. \ref{result}).
 
Additionally, to better generate the semantic information from the images in a batch, we develop two simple yet effective modules, i.e., a multi-head generator and token matcher that can catch the batch-wise semantic information. Combining them, we design a novel and sufficient method called \textit{Semantic Attention Network (SAN)} for UISS.
Our method achieves state-of-the-art performance on five different datasets in various scenarios. 

Notice that some of the compared approaches utilized additional information via leveraging pretrained models to reinforce the power of features to represent semantics. These top-down methods employed a pretrained model as the backbone and applied cluster methods on outputs of the backbone to compensate for the lack of semantic information. Employing these pretrained models brings in a massive amount of prior knowledge, which can be attributed to another question as to how to map prior knowledge to each pixel. Though our method may be unfair to compete with these methods, it still outperforms them in most cases.

In summary, our contributions are four-fold:

\begin{itemize}
    \item We propose the core properties of how to build a pixel-wise UISS model, in which we present a connection and comparison between UISS and image-wise representation learning.
    
    \item We reveal and provide proof that existing MI-based methods in UISS can suffer from representation collapse.
    
    \item We introduce a robust structural network \textit{SAN} for UISS, in which the image reconstruction task is utilized to solve the representation collapse problem, and an interactive module called \textit{Semantic Attention(SEAT)} is proposed to generate pixel-wise and semantic features better.

    \item Our new approach outperforms the previous methods and is competitive with those using additional data on five challenging datasets: COCO-Stuff-3, COCO-Stuff-15, COCO-Stuff-27, Cityscapes, and POTSDAM.
\end{itemize}

\section{Related Work}

\subsubsection{Unsupervised Semantic Segmentation.}
Methods \cite{huang2022simultaneous, huang2019self} that learn the segmentation masks entirely from data with no supervision can be categorized as follows: (1) Clustering methods\cite{hwang2019segsort} consists of a two-step process. Firstly, a CNN network produces the features, which are then grouped into clusters using spherical K-means. Secondly, the CNN is trained for better feature extraction to discriminate the clusters. The model we propose is an end-to-end method that is simple for training and inference. (2)  Mutual Information maximization methods are common to use in unsupervised learning, especially for unsupervised methods based on representation learning\cite{hjelm2018learning}\cite{ji2019invariant}; the training objective is to maximize a lower bound of MI over continuous random variables between distinct views of the inputs. However, the lower bound of MI is hard to estimate, and the image-wise MI maximization is insufficient for pixel-level image segmentation. Our method avoids using MI-driven contrastive learning to directly learn the pixel-wise representation, which can deal with segmentation tasks much more effectively. (3) Pretrained model methods\cite{cho2021picie}\cite{yin2021transfgu} utilize pretrained models to generate better representation and introduce high-level semantic information to the model while they still focus on image-wise information and use additional information compared to our method.
Our proposed method is more robust and flexible for multiple class segmenting regions.

\subsubsection{Self-attention Models and Image Semantic Segmentation.}
Transformer and self-attention models \cite{zeng2022simple, zeng2023substructure} have revolutionized deep learning \cite{wang2022mvsnet, wang2023flora}. While for image segmentation, ViT maps attention among patches, which cannot easily generate fine-grained segmentation. 
It is noted that there is a recent line of efforts for \textit{supervised} image segmentation using mask classification\cite{he2017mask}\cite{carion2020end}. This paradigm first predicts a set of global mask semantic tokens, each representing a single class. Then, all pixel-wise tokens are matched with each global mask token to generate a class probability map. Maskformer\cite{cheng2021per} formulates semantic segmentation as a mask classification problem by transforming mask classification outputs into task-dependent prediction formats. Then, based on Maskformer, Panoptic Segformer\cite{li2021panoptic} and Mask2former\cite{cheng2021masked} combined multi-scale information in the backbone to generate semantic tokens and achieve the state-of-the-art result in supervised semantic segmentation. While for \textit{unsupervised} image segmentation, to the best of our knowledge, we are the first to propose a framework translating ViT as the semantic embeddings extractor and examine the effectiveness of ViT in unsupervised semantic segmentation.

\section{Methodology}
Our method aims to learn partition-friendly representations by mapping and matching pixel-wise and semantic features in a suitable latent space. We first introduce the interactive module SEAT and then analyze how to design the supervisions for UISS by revealing how MI-based methods failed in UISS. By this, we emphasize the advantages of utilizing image reconstruction instead of MI as the supervision. Based on the analysis, we design a novel approach for UISS.
Note that our method uses neither annotated information nor pretrained models while achieving a competitive performance compared to those initialized with pretrained models. 

\subsection{Semantic Attention(SEAT)}
To better connect pixel-wise and semantic features, we design a module to dynamically build relationships between pixel-wise and batch-wise semantic embeddings via the attention mechanism.

The attention mechanism calculates the weights for each value vector, which are computed by a compatibility function of the
queries with the corresponding keys. The most commonly used attention mechanism is self-attention\cite{vaswani2017attention}, which is defined as:
\begin{equation}
    Attention(Q, K, V) = softmax(\frac{QK^T}{\sqrt{d_K}})V
\end{equation}
where $\sqrt{d_K}$is the scaling factor. $Q$ is the query, $K$ is the key and $V$ is the value and they are defined as:
\begin{equation}
    Q = F_Q(x); K = F_K(x); V = F_V(x)
\end{equation}
where $x$ is the input image and $F_Q$, $F_K$ and $F_V$ are networks.

For UISS, we regard the calculation of the similarity between $Q$ and $K$ as a mapping of each pixel into a semantic and produce a segmentation map. In this case, in our model, $Q$ is defined as the pixel-wise embedding which contains local information, and $K$ is defined as semantic embedding which contains global semantic information. In self-attention, $K$ and $V$ are calculated via different networks, while in UISS, $K$ and $V$ can be the same embedding; this semantic embedding can not only undertake the task of indexing for pixel-wise embeddings but also can be used as the extracted semantic contents to construct attention information. We define a batch of $N$ images $\{x_n\}$ where $n \in \{1...N\}$, the resolution of the original image is respectively defined as $(H \times W)$. 

Thus, the definition of semantic attention for image $n$ in a batch is denoted as:
\begin{equation}
    SEAT(Q_n, S) = softmax(\frac{Q_nS^T}{\sqrt{d_S}})S
\end{equation}
where $Q_n$ is the pixel-wise embeddings which are the representations of each pixel in image $x_n$, $S$ is the semantic embedding which are the representations of each semantic and they are defined as:
\begin{equation}
\label{no}
    Q_n = F_Q(x_n); S = F_S(\sum_{n=1}^{N}(x_n))
\end{equation}
where $F_Q$ is the pixel-wise network. $F_S$ is the batch-wise semantic network.

\subsection{Supervision for UISS}
After building the relationship between semantic embeddings and pixel-wise embeddings, we notice that, for image-wise contrastive learning, \cite{wang2020understanding} identifies two essential properties: \textit{alignment} between features from positive pairs and \textit{uniformity} of the induced distribution of the (normalized) features on the hypersphere for contrastive representation learning. In UISS, for semantic embeddings, the \textit{uniformity} property should also be satisfied to better cluster pixel-wise embeddings. But for the \textit{alignment} property, as contrastive learning only focus on positive and negative pairs, UISS requires multi-class clustering, \textit{alignment} operation in UISS turns to build \textit{alignment} among features from semantic clusters. 
Therefore, partly different from image-wise contrastive learning, we propose two critical supervisions for UISS:

(1) Each semantic embedding should represent a cluster of pixel-wise embeddings.

(2) The semantic embeddings should be uniform in the vector space.

For supervision (1), as $Q_nS^T$ can be seen as the probability map of segmentation, the operation of assigning semantic embeddings $S$ for the probability map is to reconstruct the representations of the whole image(i.e., if the segmentation result is correct, when we assign an embedding with the semantic at the pixel location provided by the correct result, we synthesize a pixel-to-pixel semantic map of the original image). The synthesized semantic map should be similar to the pixel-wise embedding map $Q_n$, for both of them ideally represent the correct mapping of different semantics. Thus, we can build the \textit{alignment} supervision based on SEAT, which can be defined as;
\begin{equation}
\label{equ1}
    softmax(\frac{Q_nS^T}{\sqrt{d_S}})S \simeq Q_n
\end{equation}

For supervision (2), to better separate the pixel-wise features, we have to guarantee the uniformity of semantic embeddings via this supervision. The ideal condition is to force the embeddings to be orthogonal to each other in the vector space, which means the semantic embeddings matrix $S$ should be orthogonal. Thus, the matrix $S$ should have the following property:
\begin{equation}
\label{equ2}
    S^TS = E
\end{equation}
where $E$ is the identity matrix. 
Thus, the loss term of the model should be:
\begin{equation}
\label{equ3}
    \mathbb{L}(Q_n, S) = d(softmax(\frac{Q_nS^T}{\sqrt{d_S}})S, Q_n) + CE(S^TS - E)
\end{equation}
where $d(\cdot, \cdot)$ indicates the discrepancy between the two inputs and $CE(\cdot, \cdot)$ indicates the cross-entropy loss.
Previous MI-based methods utilize the MI constraint to calculate the discrepancy, which is easy to collapse. We argue that it is inappropriate to address this issue simply.

Different from image-wise representation learning, for unsupervised dense prediction like UISS, other than learning the \textit{uniformity} of semantic embeddings, the \textit{uniformity} of pixel-wise embeddings should also be considered. Ignoring this constraint is the critical reason for the model collapse when utilizing MI-based methods in UISS. This can be reflected in Eq. \ref{equ3}: even though the semantic embeddings $S$ satisfy the uniformity, if no supervision is imposed on $Q_n$, the model still can not guarantee the uniformity of pixel-wise embeddings and tends to collapse. Thus, we shall force two kinds of embeddings to satisfy the uniformity property.

To solve the problem, we design a simple but efficient strategy. Notice that, in UISS, there is a kind of hidden distribution that may be ignored, the pixel distribution of a single image which can be seen as a more uniformed and nature distribution to map the pixel-wise embeddings. In this case, we chose the image reconstruction task as the constraint to satisfy the uniformity, which is defined as:
\begin{equation}
\label{equ9}
    \mathbb{L}(Q_n; F_G) =  d(F_G(Q_n), x_n)
\end{equation}
where $x_n$ is the $n^{th}$ image in the batch and $Q_n$ is the pixel-wise embeddings of $x_n$, $F_G$ is the reconstruction network. 

Thus, based on Eq. \ref{no} and Eq. \ref{equ9},  Eq. \ref{equ3} turns into:
\begin{equation}
\label{equ10}
    \begin{aligned}
        & \mathbb{L}(x; F_Q, F_S, F_G) \\
        &=  d(F_G(Q_n), F_G(softmax(\frac{Q_nS^T}{\sqrt{d_S}})S), x_n)  \\
        &+ d(S^TS - E) \\
    \end{aligned}
\end{equation}
where $Q_n = F_Q(x_n)$, $S = F_S(\sum_{n=1}^{N}(x_n))$, $d(\cdot, \cdot)$ indicates the discrepancy between the inputs.

In a nutshell, The key differences between unsupervised dense representation learning and image-wise representation learning are summarized in Table. \ref{tab:differ}.

\begin{table*}[]
    \centering
    \begin{tabular}{l|c|c}
    \hline
 & \textbf{Image-wise representation learning}  & \textbf{Pixel-wise representation learning}\\
    \hline
    \textbf{Alignment}  & {Image-Image feature alignment}& {Pixel-Semantic feature alignment }  \\
    \hline
    \textbf{Uniformity}   & {Among semantic embeddings}  &  {Among semantic embeddings \& Among pixel-wise embeddings} \\
    \hline
    \end{tabular}
    \caption{Discrepancies between image-wise and pixel-wise representation learning(from contrastive learning view).}
    \label{tab:differ}
\end{table*}

\section{Overall Framework}
\begin{figure*}[t]
\centering
\includegraphics[width=0.7\textwidth]{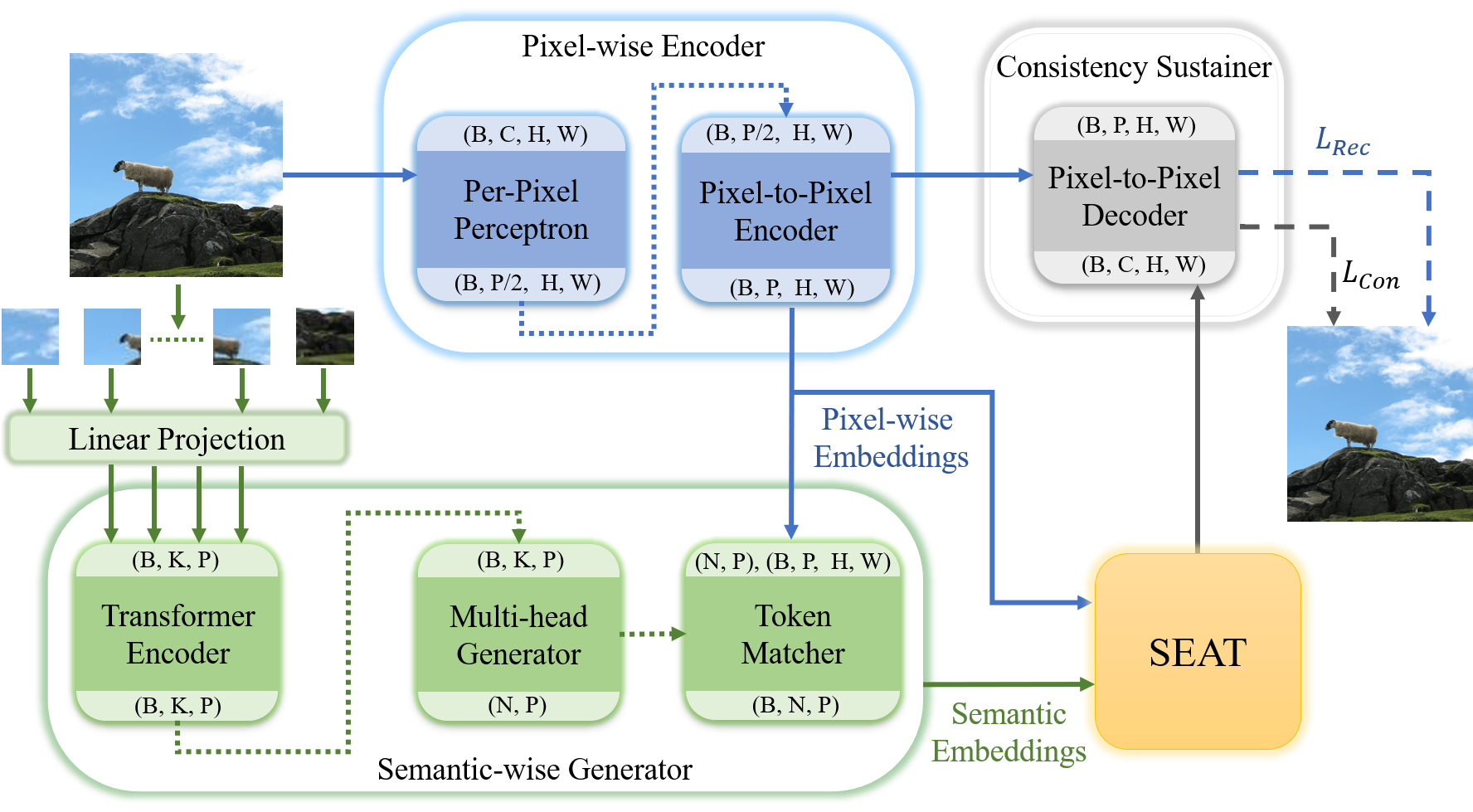} 
\caption{{ The architecture of SAN Network.} The pixel-wise encoder maps the image into a high dimensional latent space and produces pixel-wise embeddings to be clustered. The semantic-wise Generator generates semantic embeddings, serving as centers to align and group the pixel-wise features. After processing by the SEAT module, the consistency sustainer maintains the consistency between the pixel-wise features (from the pixel-wise encoder) and generated pixel-wise features (reconstructed from SEAT) to provide a constraint for the model learning. The $L_{Rec}$ and $L_{Con}$ are the $L_2$ loss function.}
\label{main}
\end{figure*}

Based on the analysis, we design a novel model for UISS.
Our model comprises three parts: pixel-wise encoder, semantic-wise generator, and consistency sustainer. As shown in Fig.\ref{main}, the pixel-wise encoder is convolution-based and designed to leverage the local information confined to the given receptive field and project the pixel-wise information into a high dimensional latent space. Semantic-wise generator tends to extract the batch-wise semantic global information and projects them into the corresponding latent space. Consistency sustainer mainly focuses on reinforcing the consistency between information from the pixel-wise and semantic embeddings and maintains the pixel-wise information from the pixel-wise encoder. 

\subsection{Pixel-wise Encoder}
We denote the training dataset for the unsupervised problem as $D=\{I_{H\times W}\}$, where the resolution of the original image is respectively defined as $(H, W)$. As shown in Fig.\ref{main}, the per-pixel perceptron uses a CNN to generate a representation from the adjacent receptive field for each pixel, denoted as the per-pixel region embedding, and the input dimension is scaled to $P/2$, where $P$ are defined as the length of the pixel-wise and semantic embeddings. Then the pixel-to-pixel encoder applies the per-pixel region embeddings to the standard length of $P$, so the output of the pixel-wise encoder is $(P, H ,W)$, denoted as pixel-wise embeddings $\zeta_d \in \mathbb{R}^{(P, H, W)}$, where $d \in D$, which represents the adjacent region descriptor for each pixel in the image.

\subsection{Semantic-wise Generator}
Previous self-supervised CNN-based approaches\cite{zhang2016colorful}\cite{pathak2016context}\cite{chen2020generative} arrange different pretext tasks for specially designed models to generate semantic-wise semantic representation.
However, the image is a higher dimensional, noisier, and more redundant modality than text\cite{he2022masked}. When a model uses unsupervised CNN structures, the inherent defects of CNN make it hard to focus on global semantic information instead of local information like texture because of no supervision. Recent work\cite{naseer2021intriguing} showed that, instead of considering low-level features like texture, ViTs perform better on shape recognition than CNNs and are comparable to humans. Therefore, we propose a ViT-based generator to generate specific semantic-wise semantic representations instead of CNN to achieve the state-of-the-art result. As far as we know, we are the first to explore unpretrained ViT in UISS, and the ablation study is presented in Table \ref{ab0}. 

For details, the transformer encoder is used to extract information from each input image by using standard transformer blocks. The MLP-based multi-head generator aims to decouple each semantic representation for each class. Notice that we remove the commonly used one layer MLP after the transformer blocks and merge all the output information, then send the information into the multi-head generator, so the output dimension of transformer blocks is $(K, P)$, where $K$ is the resulting number of patches, calculated by $(H/c) \times (W/c)$, $c$ is the resolution of each image patch. The multi-head generator remaps the batch-wise semantic information into the assigned number of semantic embeddings $\zeta_n \in \mathbb{R}^{(N, P)}$, where $N$ is the number of the semantics of the entire dataset, $n \in N$. Then giving each semantic embedding a class probability prediction $p_n \in \Delta^N$ where $\Delta^N$ is the $N$-dimensional probability simplex. For each image in dataset $D$, the embedding rematcher can choose the suitable classes of semantic embeddings. The output can be defined as $\Gamma_d = p_n \times \zeta_n $ where $d \in D$. In short, the semantic-wise generator aggregates the batch-wise information, arranges the semantics for each image, then the semantic embeddings $\Gamma_d$ and $\zeta_d$ are sent into SEAT to calculate the semantic attention, and the output of the SEAT module is called \textit{alignment embeddings} which is defined as: $\Theta_d$ These embeddings will soon be fed into the next module.

\begin{figure}[t]
\centering
\includegraphics[width=0.95\columnwidth]{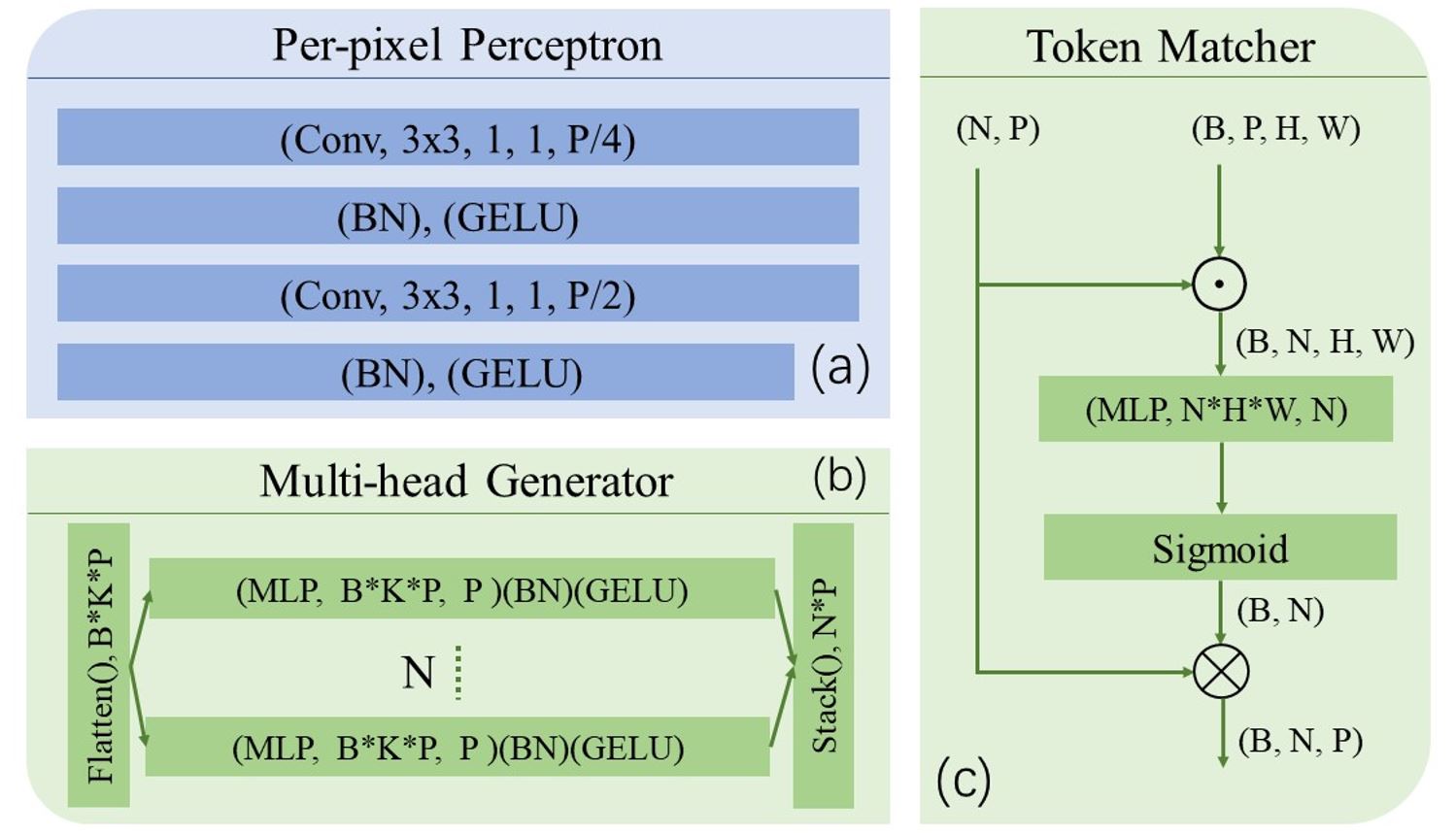} 
\caption{{ The network structures of SAN.} (a) Per-pixel Perceptron; (b) Multi-head Generator; (c) Token Matcher. { Legend}: (Conv, $K\times K$, $D$, $S$, $P$): Convolution with filter size $K\times K$, padding $D$, stride $S$ and channel $P$; (MLP, M, N): MLP with input dimension M and output dimension N; (BN): Batch Normalization; (GELU): Gaussian Error Linerar Units; (Sigmoid): Sigmoid Function.}
\label{module}
\end{figure}

\subsection{Consistency Sustainer}
\begin{figure*}[t]
\centering
\includegraphics[width=0.7\textwidth]{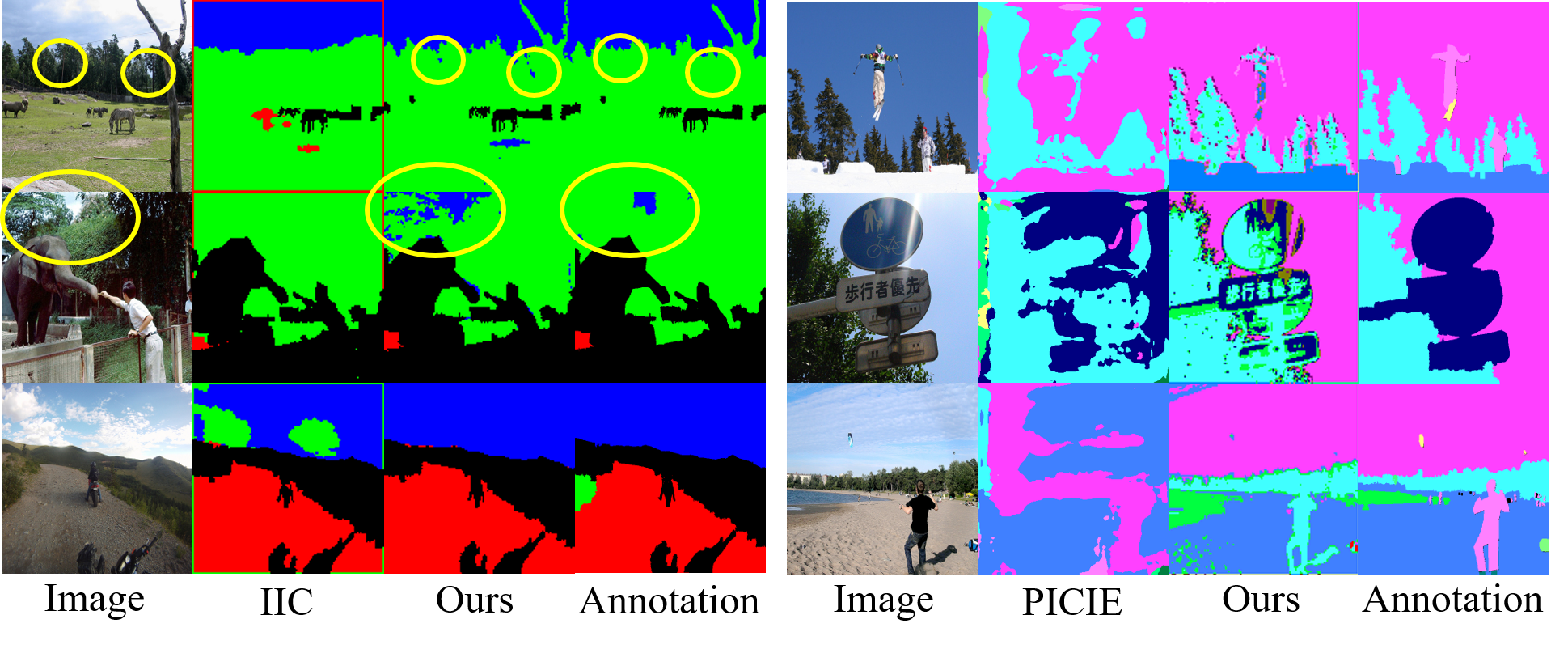} 
 \caption{{ Qualitative comparison of SAN.} { Left: Results on COCO-Stuff-3}, for this dataset, we compare our method with IIC\cite{ji2019invariant} which uses cluster methods. Our results even label more fine-grained annotations than ground truth(Line 2); { Right: Results on COCO-Stuff-27}, for the tougher dataset, we compare our result with PiCIE\cite{cho2021picie} which is an outstanding pretrained-model-based method.}
\label{result}
\end{figure*}

This module is designed to satisfy the supervisions in Eq. \ref{equ10}. We aim to minimize the distance between alignment embeddings $\Theta_d$, pixel-wise embeddings $\zeta_d$, and the input image $x_d$. We simply design a shallow $1\times 1$ convolution neural network to reconstruct the input image. The alignment embeddings $\Theta_d$ and pixel-wise embeddings $\zeta_d$ share the same network to satisfy the first loss term together in Eq. \ref{equ10}. The network can ensure the network finds the best solution to the alignment of the semantic-analogous pixel-wise embeddings. Moreover, the image reconstruction task makes the model easy to learn structure information. The visualization is presented in Fig. \ref{result}. 

For the last loss term of Eq. \ref{equ10}, we add a cosine similarity loss to force the semantic embeddings to get more uniform. While we find the result does not improve, we think this is because the normalization layer in the network can naturally orthogonalize the embeddings.

\section{Experiments}

The experiments are conducted on five semantic segmentation benchmarks of different scenarios, including COCO-Stuff-3\cite{caesar2018coco}, COCO-Stuff-15, COCO-Stuff-27, Cityscapes\cite{Cordts2016Cityscapes} and POTSDAM\cite{rottensteiner2014isprs} with various segmentation targets. COCO-Stuff-15 has 15 coarse labels, and the images are reduced to 52k by taking only images with at least 75\% stuff pixels; COCO-Stuff-3 is labeled only sky, ground, and plants. Another dataset we use is POTSDAM, which is divided into 8550 RGBIR 200 × 200 px satellite images, of which 3150 are unlabelled. There are 6-label variants (roads and cars, vegetation and trees, buildings, and clutter). We followed \cite{ji2019invariant} to build the above dataset. To show the superiority, we further tested our model on more challenging datasets and compared it with methods that are initialized with pretrained weights. We applied COCO-stuff-27, with all categories merged into 27 superclasses as \cite{cho2021picie}. We also apply the widely used Cityscapes dataset, which contains 5,000 images focusing on street scenes, divided into 2,975 and 500 images used for training and validation. All pixels are categorized into 34 classes. We follow \cite{cho2021picie} to merge 34 classes into 27 classes after filtering out the ‘void’ class. we follow the standard evaluation protocols for unsupervised segmentation used in \cite{cho2021picie}\cite{ji2019invariant}, Hungarian matching algorithm\cite{kuhn1955hungarian} is used to align indices between predictions and the ground-truth labels over all the images in the validation set. Pixel Accuracy(Acc) is reported over all the semantic categories.

For network structure, the proposed six layers ViT block is the simplest transformer mentioned in \cite{vaswani2017attention} with a random initialization; the input patch size is set to $8\times8$ and images are cropped to $128\times128$. The length of all of the embeddings is set to 1024. All datasets are trained for 30 epochs on 4 $\times$ Tesla V100 32G GPUs with Adam optimizer. The learning rate is set to $1e-4$.
For the POTSDAM dataset, we applied data augmentations such as color jittering, horizontal flipping, and color dropping following \cite{ji2019invariant}.

\begin{table}[t]
    \centering
    \begin{tabular}{l|ccccc}
    \hline
    Methods  & w/o ad. & COCO3 & POTS.  & COCO15\\
    \hline
    Isola         & $\times$  & 54.0 & 44.9 & - \\
    DeepClu.   & $\times$  & 41.6 & 29.2 & 19.9 \\
    MDC        & $\times$ & -& -& 25.3 \\
    IIC        & $\times$  & 72.3 & 45.4 & 27.9 \\
    AC         & $\times$  & 72.9 & 49.3 & 30.8 \\
    InMARS     & $\times$  & 73.1 & 47.3 & 31.0\\
    Infoseg  & $\times$  & 73.8 & 57.3 & 38.8 \\
    $\rm FS^4$        & $\times$ &- & - & 48.5  \\
    \hline
    SAN  & $\times$      & \textbf{80.3} & \textbf{60.5} & \textbf{55.7}\\
    \hline
    \end{tabular}
    \caption{{ COCO-Stuff-3/ POTSDAM/ COCO-Stuff-15.} Comparison with unpretrained SOTA methods. \textit{Add. means additional data.}}
    \label{coco3}
\end{table}

\begin{table}[t]
    \centering
    \begin{tabular}{l|cccc}
    \hline
Methods & w/o addi. & COCO-27 & Cityscapes \\
\hline
IIC & \checkmark & 21.0  & 29.8  \\
MoCoV2 & -           & 25.2 & 33.1 \\
DINO & -         & 30.5  & 37.9    \\
MDC & \checkmark  & 32.2  & 40.7  \\
$\rm FS^4$ & \checkmark & -  &41.7 \\
PiCIE(RN18) & \checkmark  & 34.2 & -\\
PiCIE(RN50) & \checkmark & 48.1 & -  \\
PiCIE+H.& \checkmark & 50.0 & \textbf{65.5}\\
\hline
SAN (ours)    & $\times$     & {\bf 52.0}(+2.0)   & 51.0(-14.5)  \\

    \hline
    \end{tabular}
    \caption{{ Cityscapes/ COCO-stuff-27.} Comparison with pretrained SOTA methods. \textit{Add. means additional data. RN18 and RN50 means ResNet18 and ResNet50 pretrained model. (+H.) means the model uses auxiliary clustering.}}
    \label{cityscapes}
\end{table}

\subsection{Main Results}

In this section, we report our results and compare them to state-of-the-art methods on the five datasets. Both pretrained and unpretrained methods are compared. We qualitatively visualize the result in Fig. \ref{result}. SAN catches more region and pixel-wise information than previous state-of-the-art methods, which is effective for unsupervised segmentation. Some results are even more accurate than ground-truth annotations! For example, when segmenting (highlighted in the yellow circles in Fig.\ref{result}), it is evident that our segmentation results are more meaningful than the ground truth.

We first test our methods on traditionally used datasets in UISS, including COCO-Stuff-3, COCO-Stuff-15, and POTSDAM. The compared methods are \cite{ji2019invariant}, \cite{ouali2020autoregressive}, \cite{mirsadeghi2021unsupervised} and \cite{wang2022fully} etc.  Table~\ref{coco3} shows the results of the comparisons. CTAE obtains impressive results on these datasets, particularly outperforming Infoseg\cite{harb2021infoseg} with a relative performance gain of 8.8\% on COCO-Stuff-3, 5.6\% on POTSDAM, and 14.9\%  on COCO-Stuff-15. 

For COCO-Stuff-27, only pretrained methods\cite{chen2020improved}, \cite{caron2021emerging}, \cite{cho2021picie} are trained and tested on this dataset; in this case, our unpretrained method still surpasses the pretrained methods. We further test our model on Cityscapes. This dataset has much fewer images(2975) compared to COCO-Stuff(50000+), which is a severe negative impact on our model, which does not use additional data. Our model still outperforms the pretrained-model-based methods\cite{ji2019invariant} .

\begin{table}[t]
    \centering
    \begin{tabular}{l|ccccc}
    \hline
     & ResNet18 & UNet & 3x3 Conv  & ViT-4& ViT-6\\
\hline
Enc. & 73.5 & 71.9  & \textbf{80.3}&- &-  \\
\hline
Dec.  & 77.8 & 78.5  & - & 79.9&\textbf{80.3}  \\

    \hline
    \end{tabular}
    \caption{Ablation study on backbones. Each backbone is replaced on the basis of the original network. \textit{Enc. means encoder, Dec. means decoder.}}
    \label{ab0}
\end{table}

\begin{table}[t]
    \centering
    \begin{tabular}{cccc}
    \hline
    Image recon. &Batch Wise & Token Matcher & Acc.(\%) \\
    \hline
     \em& \checkmark& \checkmark & 34.2\\
     \checkmark&\em & \em & 74.4 \\
    \checkmark&\em & \checkmark & 79.2 \\
    \checkmark&\checkmark & \em & 76.7 \\
    \checkmark&\checkmark & \checkmark & {\bf 80.3} \\
    \hline
    \end{tabular}
    \caption{Effect of Image Reconstruction, Batch-wise Info. and Token Matcher.}
    \label{ab1}
\end{table}

\subsection{Ablation Study}
\textbf{Pixel-wise Encoder and Semantic-wise Generator.}
We first tested the effects of different backbones on the pixel-wise encoder and semantic-wise generator. The results are presented in Table \ref{ab0}. For the pixel-wise encoder, we utilized ResNet and UNet as comparisons. The deeper neural networks of ResNet and UNet perform much worse than the shallow neural network using a superficial 3x3 Conv layer. Bear in mind that a pixel-wise generator aims to generate local-pixel information rather than wider scale batch-wise semantic information. Therefore, a simple shallow network would meet this purpose. While ResNet and UNet with deeper layers, on the contrary, focus on the whole picture of the image with a sequence of convolutions. For the semantic-wise generator, we can see that all the deep networks perform well, with deeper ViT performing the best. It might be because, compared to CNN-based models, ViT catches wider-scale semantic information than CNN through the correlation between multiple patches, better aggregating information from different regions of an image. 

\textbf{Effect of Image Reconstruction, Batch-wise Info. and Token Matcher.}
In Fig. \ref{module}, we can see that if we move the image reconstruction task, the model collapsed with a very poor result.
It is noted that we introduced batch-wise information for the multi-head generator and the token matcher on the semantic-wise generator, as shown in Fig. \ref{module}.  
We further test their effects; the results are shown in Table \ref{ab1}. Both strategies lead to significant improvements. Without batch-wise information from the multi-head generator, the performance drops by 1.1\%; without the token matcher, the performance decreases from 80.3\% to 76.7\%. With these two modules, our model performs the best. 

\section{Conclusion}
In this work, we present an in-depth analysis for UISS, in which we propose the core properties of how to build a UISS model. We also present connections and comparisons between UISS and image-wise representation learning and figure out that MI-based methods may easily collapse in UISS. Based on the above, a novel network called SAN is proposed, and a semantic attention module is proposed to generate pixel-wise and semantic features better. We hope our work can bring some fundamental enlightenment to semantic segmentation and unsupervised learning.

\section{Acknowledgments}
This work is supported in part by National Key Research and Development Program of China (2021YFF1200800) and National Natural Science Foundation of China (Grant No. 62276121).

\bibliography{aaai23}

\end{document}